%% file: caffe.tex
\documentclass{sig-alternate}
\usepackage{booktabs}
\usepackage{color}
\usepackage{tabularx}
\usepackage{rotating}
\usepackage{array}
\usepackage{hyperref}
\usepackage{tikz}
\usepackage{multirow}

\def\checkmark{\tikz\fill[scale=0.4](0,.35) -- (.25,0) -- (1,.7) -- (.25,.15) -- cycle;}
\newcommand{\chk}{\checkmark}

\begin{document}

\conferenceinfo{}{}

\title{Caffe: Convolutional Architecture \\for Fast Feature Embedding\titlenote{Corresponding Authors. The work was done while Yangqing Jia was a graduate student at Berkeley. He is currently a research scientist at Google, 1600 Amphitheater Pkwy, Mountain View, CA 94043.}}

\numberofauthors{1}

\author{
\alignauthor Yangqing Jia$^*$, Evan Shelhamer$^*$, Jeff Donahue, Sergey Karayev, \\
Jonathan Long, Ross Girshick, Sergio Guadarrama, Trevor Darrell\\
\affaddr{UC Berkeley EECS, Berkeley, CA 94702}\\
\email{\{jiayq,shelhamer,jdonahue,sergeyk,jonlong,rbg,sguada,trevor\}@eecs.berkeley.edu}
}

\maketitle
\begin{abstract}
Caffe provides multimedia scientists and practitioners with a clean and modifiable framework for state-of-the-art deep learning algorithms and a collection of reference models.
The framework is a BSD-licensed C++ library with Python and MATLAB bindings for training and deploying general-purpose convolutional neural networks and other deep models efficiently on commodity architectures.
Caffe fits industry and internet-scale media needs by CUDA GPU computation, processing over 40 million images a day on a single K40 or Titan GPU ($\approx$ 2.5 ms per image).
By separating model representation from actual implementation, Caffe allows experimentation and seamless switching among platforms for ease of development and deployment from prototyping machines to cloud environments.

Caffe is maintained and developed by the Berkeley Vision and Learning Center (BVLC) with the help of an active community of contributors on GitHub.
It powers ongoing research projects, large-scale industrial applications, and startup prototypes in vision, speech, and multimedia.
\end{abstract}

\category{I.5.1}{Pattern Recognition}[Applications--Computer vision]
\category{D.2.2}{Software Engineering}[Design Tools and Techniques--Software
libraries] \category{I.5.1}{Pattern Recognition}[Models--Neural Nets]

\terms{Algorithms, Design, Experimentation}


\section{Introduction}
A key problem in multimedia data analysis is discovery of effective representations for sensory inputs---images, soundwaves, haptics, etc.
While performance of conventional, handcrafted features has plateaued in recent years, new developments in deep compositional architectures have kept performance levels rising \cite{krizhevsky2012imagenet}.
Deep models have outperformed hand-engineered feature representations in many domains, and made learning possible in domains where engineered features were lacking entirely.

We are particularly motivated by large-scale visual recognition, where a specific type of deep architecture has achieved a commanding lead on the state-of-the-art.
These \emph{Convolutional Neural Networks}, or CNNs, are discriminatively trained via back-propagation through layers of convolutional filters and other operations such as rectification and pooling.
Following the early success of digit classification in the 90's, these models have recently surpassed all known methods for large-scale visual recognition, and have been adopted by industry heavyweights such as Google, Facebook, and Baidu for image understanding and search.


While deep neural networks have attracted enthusiastic interest within computer vision and beyond, replication of published results can involve months of work by a researcher or engineer.
Sometimes researchers deem it worthwhile to release trained models along with the paper advertising their performance.
But trained models alone are not sufficient for rapid research progress and emerging commercial applications,
and few toolboxes offer truly off-the-shelf deployment of state-of-the-art models---and those that do are often not computationally efficient and thus unsuitable for commercial deployment.

To address such problems, we present Caffe, a fully open-source framework that affords clear access to deep architectures.
The code is written in clean, efficient C++, with CUDA used for GPU computation, and nearly complete, well-supported bindings to Python/Numpy and MATLAB.
Caffe adheres to software engineering best practices, providing unit tests for correctness and experimental rigor and speed for deployment.
It is also well-suited for research use, due to the careful modularity of the code, and the clean separation of network definition (usually the novel part of deep learning research) from actual implementation.

\input{comparison_table.tex}

In Caffe, multimedia scientists and practitioners have an orderly and extensible toolkit for state-of-the-art deep learning algorithms, with reference models provided out of the box.
Fast CUDA code and GPU computation fit industry needs by achieving processing speeds of more than 40 million images per day on a single K40 or Titan GPU.
The same models can be run in CPU or GPU mode on a variety of hardware: Caffe separates the representation from the actual implementation, and seamless switching between heterogeneous platforms furthers development and deployment---Caffe can even be run in the cloud.

While Caffe was first designed for vision, it has been adopted and improved by users in speech recognition, robotics, neuroscience, and astronomy.
We hope to see this trend continue so that further sciences and industries can take advantage of deep learning.

Caffe is maintained and developed by the BVLC with the active efforts of several graduate students, and welcomes open-source contributions at \url{http://github.com/BVLC/caffe}.
We thank all of our contributors for their work!

\section{Highlights of Caffe}

Caffe provides a complete toolkit for training, testing, finetuning, and deploying models, with well-documented examples for all of these tasks.
As such, it's an ideal starting point for researchers and other developers looking to jump into state-of-the-art machine learning.
At the same time, it's likely the fastest available implementation of these algorithms, making it immediately useful for industrial deployment.

{\bfseries Modularity.}
The software is designed from the beginning to be as modular as possible, allowing easy extension to new data formats, network layers, and loss functions.
Lots of layers and loss functions are already implemented, and plentiful examples show how these are composed into trainable recognition systems for various tasks.

{\bfseries Separation of representation and implementation.}
Caffe model definitions are written as config files using the Protocol Buffer language.
Caffe supports network architectures in the form of arbitrary directed acyclic graphs.
Upon instantiation, Caffe reserves exactly as much memory as needed for the network, and abstracts from its underlying location in host or GPU.
Switching between a CPU and GPU implementation is exactly one function call.

{\bfseries Test coverage.}
Every single module in Caffe has a test, and no new code is accepted into the project without corresponding tests.
This allows rapid improvements and refactoring of the codebase, and imparts a welcome feeling of peacefulness to the researchers using the code.

{\bfseries Python and MATLAB bindings.}
For rapid prototyping and interfacing with existing research code, Caffe provides Python and MATLAB bindings.
Both languages may be used to construct networks and classify inputs.
The Python bindings also expose the solver module for easy prototyping of new training procedures.

{\bfseries Pre-trained reference models.}
Caffe provides (for academic and non-commercial use---not BSD license) reference models for visual tasks, including the landmark ``AlexNet'' ImageNet model~\cite{krizhevsky2012imagenet} with variations and the R-CNN detection model~\cite{girshick2014rcnn}.
More are scheduled for release.
We are strong proponents of reproducible research: we hope that a common software substrate will foster quick progress in the search over network architectures and applications.

\subsection{Comparison to related software}
We summarize the landscape of convolutional neural network software used in recent publications in Table \ref{tab:software}.
While our list is incomplete, we have included the toolkits that are most notable to the best of our knowledge.
Caffe differs from other contemporary CNN frameworks in two major ways:

(1) The implementation is completely C++ based, which eases integration into existing C++ systems and interfaces common in industry. The CPU mode removes the barrier of specialized hardware for deployment and experiments once a model is trained.

(2) Reference models are provided off-the-shelf for quick experimentation with state-of-the-art results, without the need for costly re-learning.
By finetuning for related tasks, such as those explored by \cite{DonahueJVHZTD13}, these models provide a warm-start to new research and applications.
Crucially, we publish not only the trained models but also the recipes and code to reproduce them.

\begin{figure*}
    \centering
    \includegraphics[width=0.8\textwidth]{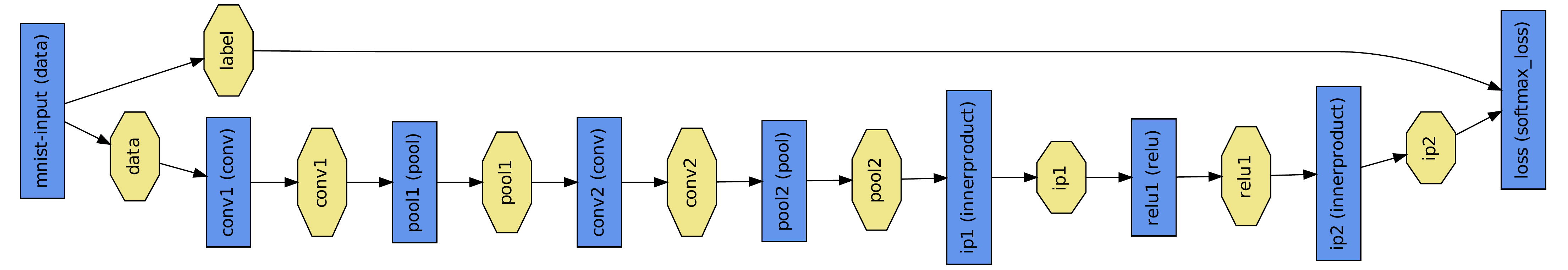}\vspace{-0.15in}
    \caption{An MNIST digit classification example of a Caffe network, where blue boxes represent layers and yellow octagons represent data blobs produced by or fed into the layers.}\label{fig:lenet}
\end{figure*}

\section{Architecture}

\subsection{Data Storage}
Caffe stores and communicates data in 4-dimensional arrays called \emph{blobs}.

Blobs provide a unified memory interface, holding batches of images (or other data), parameters, or parameter updates.
Blobs conceal the computational and mental overhead of mixed CPU/GPU operation by synchronizing from the CPU host to the GPU device as needed.
In practice, one loads data from the disk to a blob in CPU code, calls a CUDA kernel to do GPU computation, and ferries the blob off to the next layer, ignoring low-level details while maintaining a high level of performance.
Memory on the host and device is allocated on demand (lazily) for efficient memory usage.

Models are saved to disk as Google Protocol Buffers\footnote{\url{https://code.google.com/p/protobuf/}}, which have several important features: minimal-size binary strings when serialized, efficient serialization, a human-readable text format compatible with the binary version, and efficient interface implementations in multiple languages, most notably C++ and Python.

Large-scale data is stored in LevelDB\footnote{\url{https://code.google.com/p/leveldb/}} databases. In our test program, LevelDB and Protocol Buffers provide a throughput of 150MB/s on commodity machines with minimal CPU impact.
Thanks to layer-wise design (discussed below) and code modularity, we have recently added support for other data sources, including some contributed by the open source community.

\subsection{Layers}
A Caffe layer is the essence of a neural network layer: it takes one or more blobs as input, and yields one or more blobs as output.
Layers have two key responsibilities for the operation of the network as a whole: a \emph{forward pass} that takes the inputs and produces the outputs, and a \emph{backward pass} that takes the gradient with respect to the output, and computes the gradients with respect to the parameters and to the inputs, which are in turn back-propagated to earlier layers.

Caffe provides a complete set of layer types including: convolution, pooling, inner products, nonlinearities like rectified linear and logistic, local response normalization, element-wise operations, and losses like softmax and hinge.
These are all the types needed for state-of-the-art visual tasks.
Coding custom layers requires minimal effort due to the compositional construction of networks.

\subsection{Networks and Run Mode}
Caffe does all the bookkeeping for any directed acyclic graph of layers, ensuring correctness of the forward and backward passes.
Caffe models are end-to-end machine learning systems.
A typical network begins with a data layer that loads from disk and ends with a loss layer that computes the objective for a task such as classification or reconstruction.

The network is run on CPU or GPU by setting a single switch.
Layers come with corresponding CPU and GPU routines that produce identical results (with tests to prove it).
The CPU/GPU switch is seamless and independent of the model definition.

\subsection{Training A Network}
Caffe trains models by the fast and standard stochastic gradient descent algorithm.
Figure \ref{fig:lenet} shows a typical example of a Caffe network (for MNIST digit classification) during training: a data layer fetches the images and labels from disk, passes it through multiple layers such as convolution, pooling and rectified linear transforms, and feeds the final prediction into a classification loss layer that produces the loss and gradients which train the whole network. This example is found in the Caffe source code at {\tt examples/lenet/lenet\_train.prototxt}.
Data are processed in mini-batches that pass through the network sequentially.
Vital to training are learning rate decay schedules, momentum, and snapshots for stopping and resuming, all of which are implemented and documented.

Finetuning, the adaptation of an existing model to new architectures or data, is a standard method in Caffe. From a snapshot of an existing network and a model definition for the new network, Caffe finetunes the old model weights for the new task and initializes new weights as needed.
This capability is essential for tasks such as knowledge transfer \cite{DonahueJVHZTD13}, object detection \cite{girshick2014rcnn}, and object retrieval \cite{GuadarramaSEKNRJT14}.

\section{Applications and Examples}

\begin{figure}[t]
    \centering
    \includegraphics[width=0.8\linewidth]{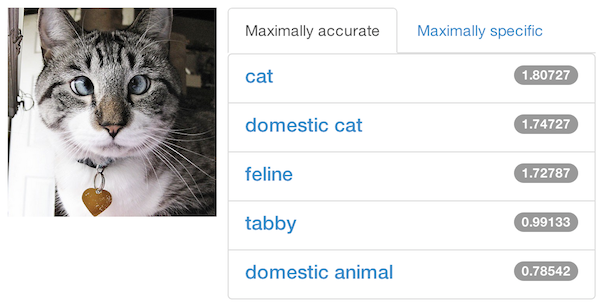}
    \caption{An example of the Caffe object classification demo. Try it out yourself online!}
    \label{fig:demo}
\end{figure}

In its first six months since public release, Caffe has already been used in a large number of research projects at UC Berkeley and other universities, achieving state-of-the-art performance on a number of tasks.
Members of Berkeley EECS have also collaborated with several industry partners such as Facebook \cite{ZhangPRDB13} and Adobe \cite{KarayevHWAD13}, using Caffe or its direct precursor (Decaf) to obtain state-of-the-art results.

{\bfseries Object Classification}
Caffe has an online demo\footnote{\url{http://demo.caffe.berkeleyvision.org/}} showing state-of-the-art object classification  on images provided by the users, including via mobile phone.
The demo takes the image and tries to categorize it into one of the 1,000 ImageNet categories\footnote{\url{http://www.image-net.org/challenges/LSVRC/2013/}}.
A typical classification result is shown in Figure \ref{fig:demo}.

Furthermore, we have successfully trained a model with all 10,000 categories of the full ImageNet dataset by fine-tuning this network.
The resulting network has been applied to open vocabulary object retrieval \cite{GuadarramaSEKNRJT14}.

\begin{figure}[t]
    \centering
    \includegraphics[width=0.25\textwidth]{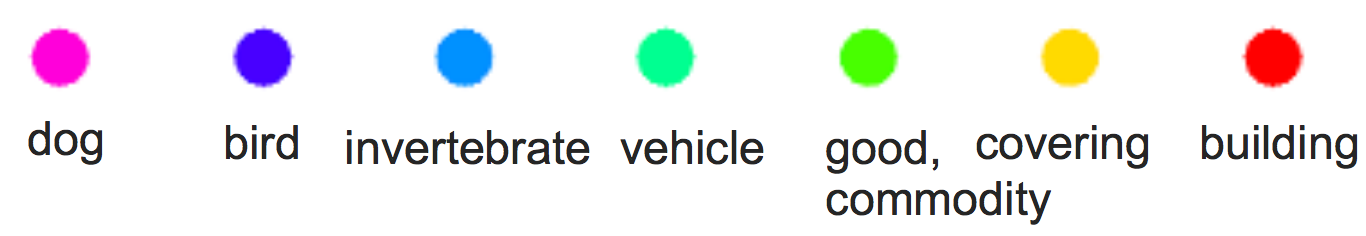}
    \includegraphics[width=0.25\textwidth]{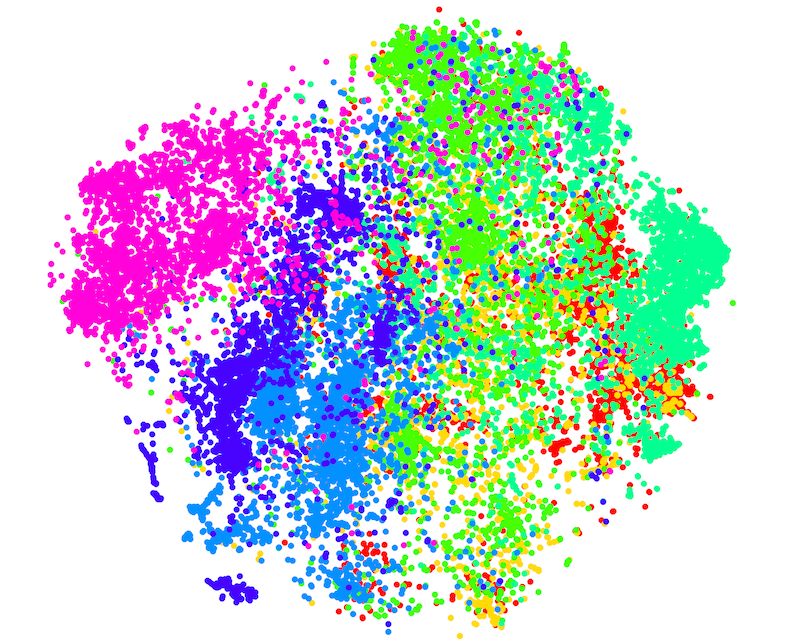}
    \caption{
    Features extracted from a deep network, visualized in a 2-dimensional space.
    Note the clear separation between categories, indicative of a successful embedding.
    }
    \label{fig:embedding}
\end{figure}

{\bfseries Learning Semantic Features}
In addition to end-to-end training, Caffe can also be used to extract semantic features from images using a pre-trained network.
These features can be used ``downstream'' in other vision tasks with great success \cite{DonahueJVHZTD13}.
Figure \ref{fig:embedding} shows a two-dimensional embedding of all the ImageNet validation images, colored by a coarse category that they come from.
The nice separation testifies to a successful semantic embedding.

Intriguingly, this learned feature is useful for a lot more than object categories.
For example, Karayev et al.\ have shown promising results finding images of different styles such as ``Vintage'' and ``Romantic'' using Caffe features (Figure \ref{fig:style}) \cite{KarayevHWAD13}.

\input{flickr_on_flickr_mini}

{\bfseries Object Detection}
Most notably, Caffe has enabled us to obtain \textbf{by far} the best performance on object detection, evaluated on the hardest academic datasets: the PASCAL VOC 2007-2012 and the ImageNet 2013 Detection challenge \cite{girshick2014rcnn}.

\begin{figure}
    \centering
    \includegraphics[width=0.45\textwidth]{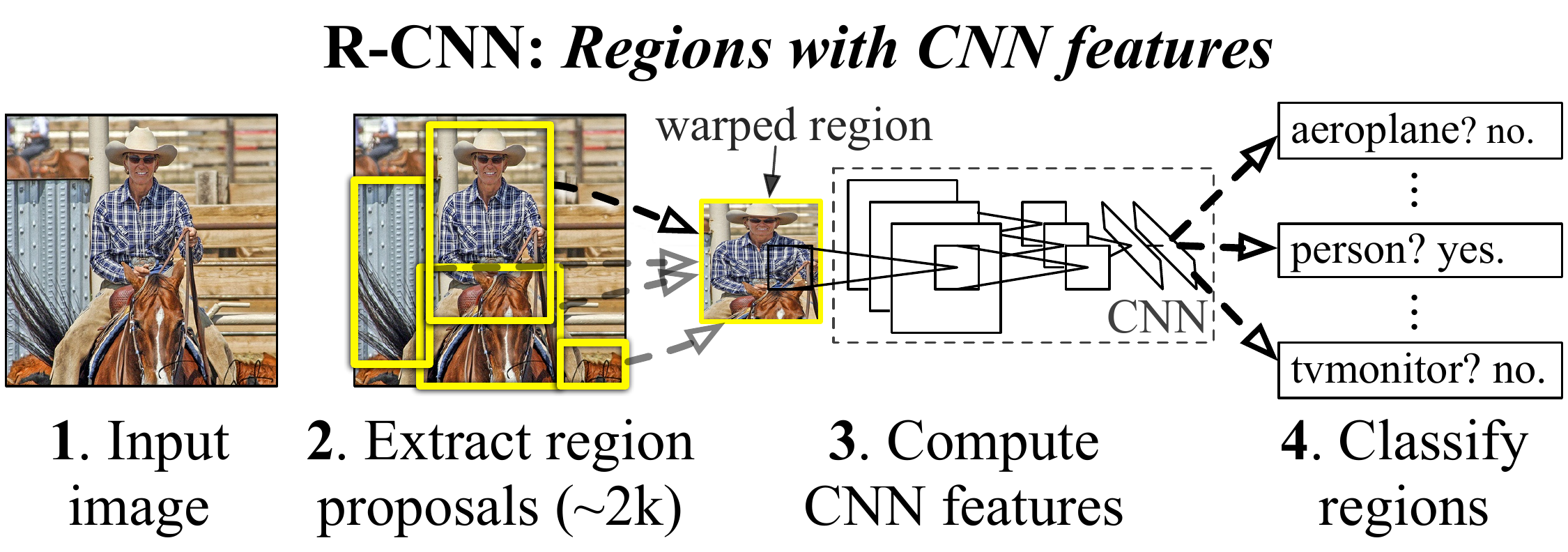}
    \caption{The R-CNN pipeline that uses Caffe for object detection.}
    \label{fig:rcnn}
\end{figure}

Girshick et al.\ \cite{girshick2014rcnn} have combined Caffe together with techniques such as Selective Search \cite{UijlingsIJCV2013} to effectively perform simultaneous localization and recognition in natural images.
Figure \ref{fig:rcnn} shows a sketch of their approach.

{\bfseries Beginners' Guides}
To help users get started with installing, using, and modifying Caffe, we have provided instructions and tutorials on the Caffe webpage. The tutorials range from small demos (MNIST digit recognition) to serious deployments (end-to-end learning on ImageNet).

Although these tutorials serve as effective documentation of the functionality of Caffe, the Caffe source code additionally provides detailed inline documentation on all modules.
This documentation will be exposed in a standalone web interface in the near future.

\section{Availability}
Source code is published BSD-licensed on GitHub.\footnote{\url{https://github.com/BVLC/caffe/}}
Project details, step-wise tutorials, and pre-trained models are on the homepage.\footnote{\url{http://caffe.berkeleyvision.org/}}
Development is done in Linux and OS X, and users have reported Windows builds.
A public Caffe Amazon EC2 instance is coming soon.

\section{Acknowledgements}
We would like to thank NVIDIA for GPU donation, the BVLC sponsors (\url{http://bvlc.eecs.berkeley.edu/}), and our open source community.

{\small
\bibliographystyle{plain}
\bibliography{sigproc}
}
\end{document}

%% file: comparison_table.tex
\newcommand{\rr}[1]{\multirow{2}{*}{#1}}
\newcommand{\chr}{\rr{\checkmark}}
\begin{table*}[t]
    \centering
    \small
    \begin{tabular}{l|lllllllll}
        \toprule
                        &             & Core     &            &      &      & Open   &          & Pretrained &              \\
        Framework       & License     & language & Binding(s) & CPU  & GPU  & source & Training & models     & Development  \\
        \midrule
        \rr{Caffe}                                     & \rr{BSD}    & \rr{C++} & Python, & \chr & \chr & \chr & \chr & \chr & \rr{distributed}  \\
                                                       &             &          & MATLAB  &      &      &      &      &      & \\
        cuda-convnet~\cite{krizhevsky2012cuda-convnet} & unspecified & C++      & Python  &      & \chk & \chk & \chk &      & discontinued      \\
        Decaf~\cite{DonahueJVHZTD13}                   & BSD         & Python   &         & \chk &      & \chk & \chk & \chk & discontinued      \\
        OverFeat~\cite{sermanet2014overfeat}           & unspecified & Lua      & C++,Python     & \chk &      &      &      & \chk & centralized       \\
        Theano/Pylearn2~\cite{goodfellow013pylearn2}   & BSD         & Python   &         & \chk & \chk & \chk & \chk &      & distributed       \\
        Torch7~\cite{collobert2011torch7}              & BSD         & Lua      &         & \chk & \chk & \chk & \chk &      & distributed       \\
        \bottomrule
    \end{tabular}
    \caption{Comparison of popular deep learning frameworks.
      \textit{Core language} is the main library language, while \textit{bindings} have an officially supported library interface for feature extraction, training, etc.
      \textit{CPU} indicates availability of host-only computation, no GPU usage (e.g., for cluster deployment); \textit{GPU} indicates the GPU computation capability essential for training modern CNNs.}
    \label{tab:software}
\end{table*}

%% file: flickr_on_flickr_mini.tex
\newcommand{\hide}[1]{}
\hide{
\newcommand{\fgap}{.5in}
\begin{figure}[h!]
\centering
\begin{tabular}{m{.02in}|m{\fgap} m{\fgap} m{\fgap} m{\fgap} m{\fgap}}
    \begin{turn}{90}\footnotesize{Minimal}\end{turn} &
    \includegraphics[width=.5in]{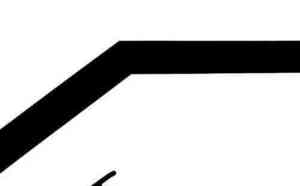} &
    \includegraphics[width=.5in]{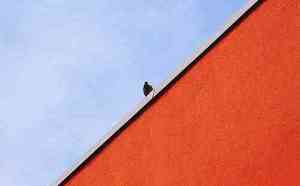} &
    \includegraphics[width=.5in]{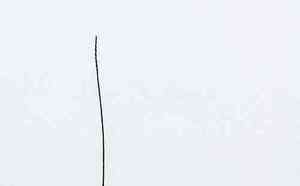} &
    \includegraphics[width=.5in]{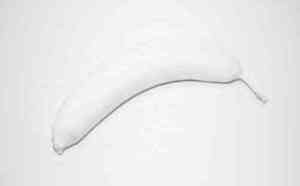} &
    \includegraphics[width=.5in]{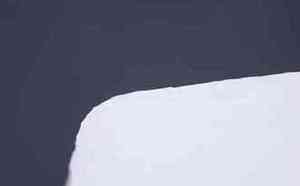} \\
    \begin{turn}{90}\footnotesize{Melancholy}\end{turn} &
    \includegraphics[width=.5in]{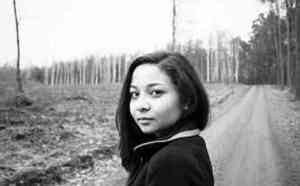} &
    \includegraphics[width=.5in]{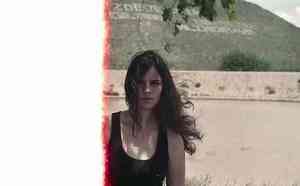} &
    \includegraphics[width=.5in]{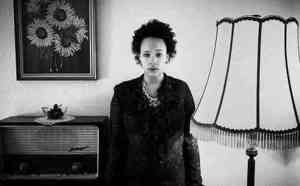} &
    \includegraphics[width=.5in]{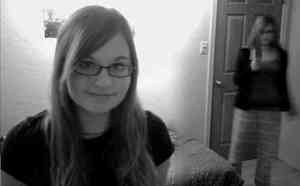} &
    \includegraphics[width=.5in]{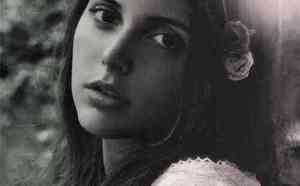} \\
    \begin{turn}{90}\footnotesize{HDR}\end{turn} &
    \includegraphics[width=.5in]{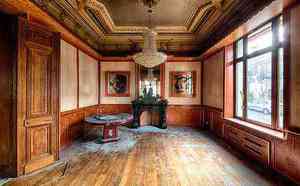} &
    \includegraphics[width=.5in]{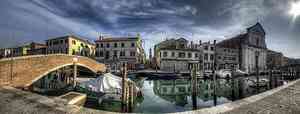} &
    \includegraphics[width=.5in]{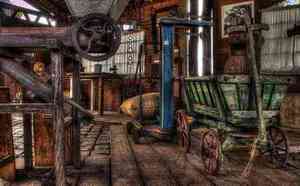} &
    \includegraphics[width=.5in]{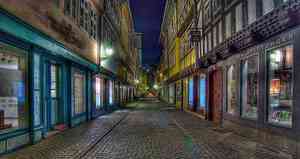} &
    \includegraphics[width=.5in]{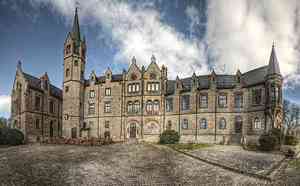} \\
    \begin{turn}{90}\footnotesize{Ethereal}\end{turn} &
    \includegraphics[width=.5in]{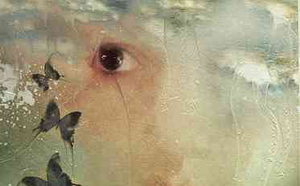} &
    \includegraphics[width=.5in]{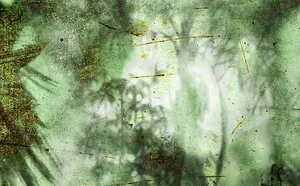} &
    \includegraphics[width=.5in]{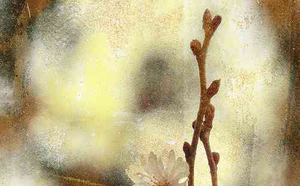} &
    \includegraphics[width=.5in]{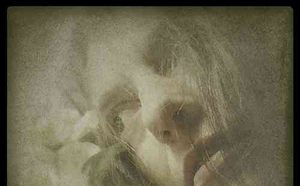} &
    \includegraphics[width=.5in]{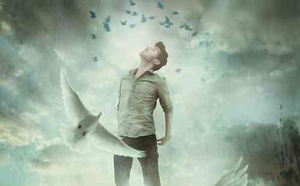} \\
\end{tabular}
\vspace{1em}
\caption{
    Top five most-confident positive predictions on the Flickr Style dataset, using a Caffe-trained classifier.
}\label{fig:style}
\end{figure}
}

\begin{figure}[h!]
\centering
\begin{tabular}{cccc}
    \toprule
    Ethereal & HDR & Melancholy & Minimal \\
    \includegraphics[width=.5in]{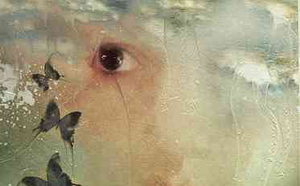} &
    \includegraphics[width=.5in]{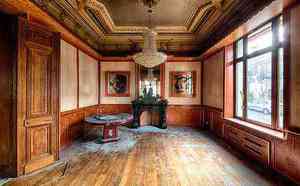} &
    \includegraphics[width=.5in]{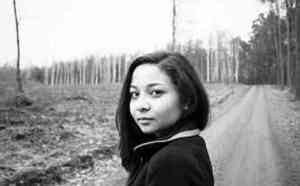} &
    \includegraphics[width=.5in]{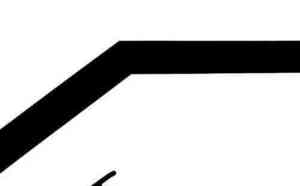} \\
    \includegraphics[width=.5in]{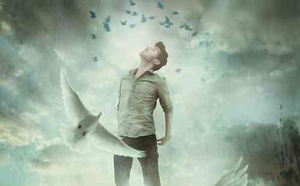} &
    \includegraphics[width=.5in]{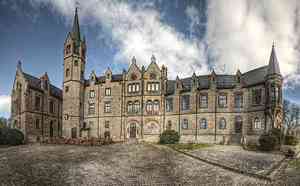} &
    \includegraphics[width=.5in]{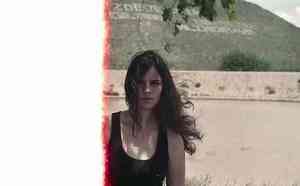} &
    \includegraphics[width=.5in]{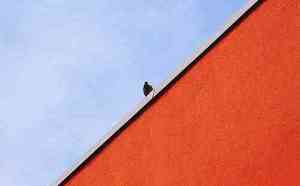} \\
    \includegraphics[width=.5in]{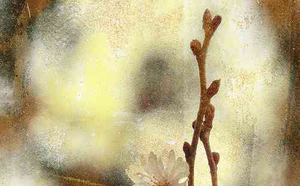} &
    \includegraphics[width=.5in]{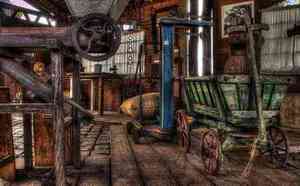} &
    \includegraphics[width=.5in]{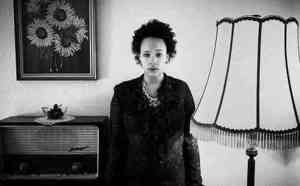} &
    \includegraphics[width=.5in]{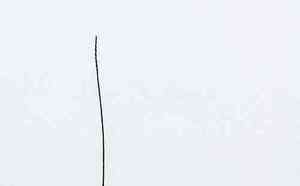} \\
    \bottomrule
\end{tabular}
\caption{
    Top three most-confident positive predictions on the Flickr Style dataset, using a Caffe-trained classifier.
}\label{fig:style}
\end{figure}

%% file: caffe.bbl
\begin{thebibliography}{10}

\bibitem{collobert2011torch7}
R.~Collobert, K.~Kavukcuoglu, and C.~Farabet.
\newblock Torch7: A {MATLAB}-like environment for machine learning.
\newblock In {\em BigLearn, NIPS Workshop}, 2011.

\bibitem{DonahueJVHZTD13}
J.~Donahue, Y.~Jia, O.~Vinyals, J.~Hoffman, N.~Zhang, E.~Tzeng, and T.~Darrell.
\newblock Decaf: A deep convolutional activation feature for generic visual
  recognition.
\newblock In {\em ICML}, 2014.

\bibitem{girshick2014rcnn}
R.~Girshick, J.~Donahue, T.~Darrell, and J.~Malik.
\newblock Rich feature hierarchies for accurate object detection and semantic
  segmentation.
\newblock In {\em CVPR}, 2014.

\bibitem{goodfellow013pylearn2}
I.~Goodfellow, D.~Warde-Farley, P.~Lamblin, V.~Dumoulin, M.~Mirza, R.~Pascanu,
  J.~Bergstra, F.~Bastien, and Y.~Bengio.
\newblock Pylearn2: a machine learning research library.
\newblock {\em arXiv preprint 1308.4214}, 2013.

\bibitem{GuadarramaSEKNRJT14}
S.~Guadarrama, E.~Rodner, K.~Saenko, N.~Zhang, R.~Farrell, J.~Donahue, and
  T.~Darrell.
\newblock Open-vocabulary object retrieval.
\newblock In {\em RSS}, 2014.

\bibitem{KarayevHWAD13}
S.~Karayev, M.~Trentacoste, H.~Han, A.~Agarwala, T.~Darrell, A.~Hertzmann, and
  H.~Winnemoeller.
\newblock Recognizing image style.
\newblock {\em arXiv preprint 1311.3715}, 2013.

\bibitem{krizhevsky2012cuda-convnet}
A.~Krizhevsky.
\newblock cuda-convnet.
\newblock https://code.google.com/p/cuda-convnet/, 2012.

\bibitem{krizhevsky2012imagenet}
A.~Krizhevsky, I.~Sutskever, and G.~Hinton.
\newblock Image{N}et classification with deep convolutional neural networks.
\newblock In {\em NIPS}, 2012.

\bibitem{sermanet2014overfeat}
P.~Sermanet, D.~Eigen, X.~Zhang, M.~Mathieu, R.~Fergus, and Y.~LeCun.
\newblock Overfeat: Integrated recognition, localization and detection using
  convolutional networks.
\newblock In {\em ICLR}, 2014.

\bibitem{UijlingsIJCV2013}
J.~Uijlings, K.~van~de Sande, T.~Gevers, and A.~Smeulders.
\newblock Selective search for object recognition.
\newblock {\em IJCV}, 2013.

\bibitem{ZhangPRDB13}
N.~Zhang, M.~Paluri, M.~Ranzato, T.~Darrell, and L.~Bourdev.
\newblock Panda: Pose aligned networks for deep attribute modeling.
\newblock In {\em CVPR}, 2014.

\end{thebibliography}
